\documentclass[5p, times]{elsarticle}
\usepackage{prletters, hyperref, float, svg, booktabs, graphicx, tabularx, subcaption,  lineno, hyperref, color, mathtools, amssymb, amsmath, algorithm2e}
\journal{Pattern Recognition Letters}
\graphicspath{ {./} }

\begin{document}

\begin{frontmatter}

    \title{Adversarial Image Generation by Spatial Transformation in Perceptual Colorspaces}

    \author[]{Ayberk \snm{Aydin}\corref{cor1}}
    \cortext[cor1]{Corresponding author:
    }
    \ead{aayberk@metu.edu.tr}
    \author[]{Alptekin \snm{Temizel}}
    \ead{atemizel@metu.edu.tr}

    \address{Graduate School of Informatics, Middle East Technical University, Turkey}


    \begin{abstract}
        Deep neural networks are known to be vulnerable to adversarial perturbations. The amount of these perturbations are generally quantified using \(\mathcal{L}_p\) metrics, such as \(\mathcal{L}_0\), \(\mathcal{L}_2\) and \(\mathcal{L}_\infty\). However, even when the measured perturbations are small, they tend to be noticeable by human observers since \(\mathcal{L}_p\) distance metrics are not representative of human perception. On the other hand, humans are less sensitive to changes in colorspace. In addition, pixel shifts in a constrained neighborhood are hard to notice. Motivated by these observations, we propose a method that creates adversarial examples by applying spatial transformations, which creates adversarial examples by changing the pixel locations independently to chrominance channels of perceptual colorspaces such as \(YC_{b}C_{r}\) and \(CIELAB\), instead of making an additive perturbation or manipulating pixel values directly. In a targeted white-box attack setting, the proposed method is able to obtain competitive fooling rates with very high confidence. The experimental evaluations show that the proposed method has favorable results in terms of approximate perceptual distance between benign and adversarially generated images. The source code is publicly available at \url{https://github.com/ayberkydn/stadv-torch}
    \end{abstract}


\end{frontmatter}
\section{Introduction}
In recent years, deep neural networks have shown impressive performance in many vision related tasks such as image classification~\cite{he2015deep}, object detection~\cite{redmon2018yolov3} and image segmentation~\cite{long2015fully}. However, they are found to be vulnerable to intentionally crafted small perturbations called \textit{adversarial perturbations}~\cite{szegedy2013intriguing}. These small perturbations added to the input image successfully change the output of a trained classifier by altering the logits large enough to change its decision to a preferred class~\cite{goodfellow2014explaining}. Adversarial examples, which can deceive machine learning models, have potential applications in distinguishing humans from algorithms. While humans can still interpret these images accurately, machine learning models can be misled by the added perturbations. For such a system to be effective, the perturbations must be small enough to mislead the algorithm while not distracting human vision. An example use case for adversarial images is Completely Automated Public Turing test to tell Computers and Humans Apart~(CAPTCHA) systems which typically involve presenting the user with an image that is easy for a human to classify, but difficult for a computer. The aim of this study is to produce successful adversarial attacks minimizing the perturbations that might interfere with human perception. 

\begin{figure}[t]
    \centering
    \begin{subfigure}[b]{.32\linewidth}
        \includegraphics[width=\linewidth]{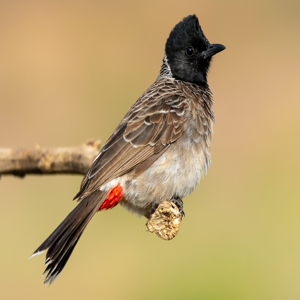}
        \caption{}
    \end{subfigure}
    \begin{subfigure}[b]{.32\linewidth}
        \includegraphics[width=\linewidth]{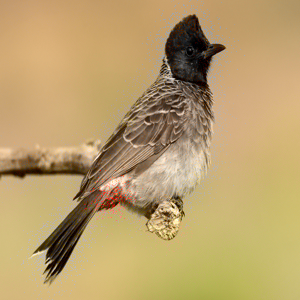}
        \caption{}
    \end{subfigure}
    \begin{subfigure}[b]{.32\linewidth}
        \includegraphics[width=\linewidth]{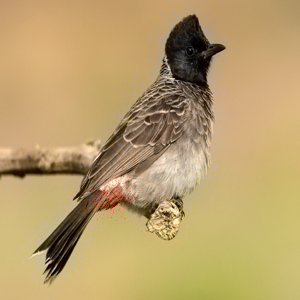}
        \caption{}
    \end{subfigure}
    \begin{subfigure}[b]{.32\linewidth}
        \includegraphics[width=\linewidth]{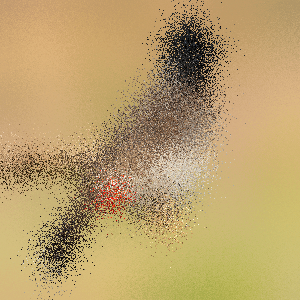}
        \caption{}
    \end{subfigure}
    \begin{subfigure}[b]{.32\linewidth}
        \includegraphics[width=\linewidth]{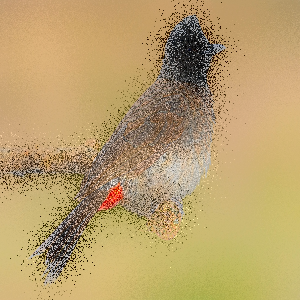}
        \caption{}
    \end{subfigure}
    \begin{subfigure}[b]{.32\linewidth}
        \includegraphics[width=\linewidth]{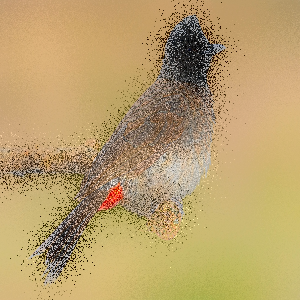}
        \caption{}
    \end{subfigure}
    \caption{Effect of flow field applied to different channels, (a) original image, Images where flow field is applied to (b) \(C_{b}C_{r}\), (c) \(a^*b^*\), (d) RGB, (e) Y and (f) L channel. The magnitude of the flow is scaled up to emphasize the effect for illustration.}\label{fig:flowtochannels}
\end{figure}

\begin{figure}[t]
    \centering
    \begin{subfigure}[b]{.23\linewidth}
        \includegraphics[width=\linewidth]{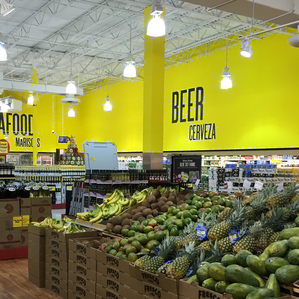}
        \caption{}
    \end{subfigure}
    \begin{subfigure}[b]{.23\linewidth}
        \includegraphics[width=\linewidth]{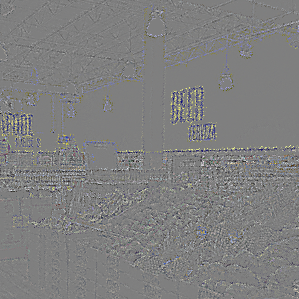}
        \caption{}
    \end{subfigure}
    \begin{subfigure}[b]{.23\linewidth}
        \includegraphics[width=\linewidth]{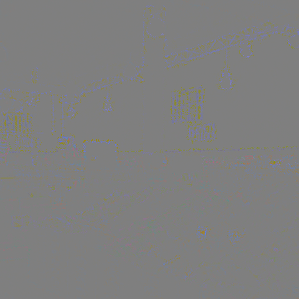}
        \caption{}
    \end{subfigure}
    \begin{subfigure}[b]{.23\linewidth}
        \includegraphics[width=\linewidth]{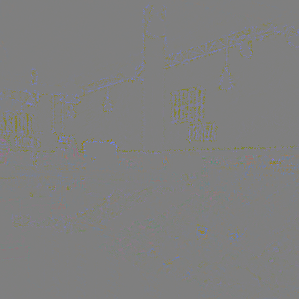}
        \caption{}
    \end{subfigure}
    \caption{Visual difference from flow field applied to different channels, (a) original image, Visualization of pixel differences where flow field is applied to (b) RGB, (c) \(C_{b}C_{r}\), (d) \(a^*b^*\) channels. The magnitude of the flow is scaled up and contrast of the pixel differences is increased to enhance the visibility for illustration. }\label{fig:diff}
\end{figure}

Adversarial perturbations are generally minimized according to \(\mathcal{L}_p\) metrics~\cite{carlini2017towards} such as \(\mathcal{L}_0\), corresponding to the number of pixels perturbed~\cite{su2019one, keppel2022explainable}; \(\mathcal{L}_2\), euclidean distance between the benign and adversarial image~\cite{carlini2017towards} or \(\mathcal{L}_\infty\), maximum magnitude of perturbation among all the pixels~\cite{goodfellow2014explaining}. However, they are generally visible to human observers, since a small \(\mathcal{L}_p\) does not always correspond to small visible perturbations~\cite{jordan2019quantifying,engstrom2018rotation}.
There is an ongoing research on finding difference metrics over 2D images and color spaces that aligns with human visual perception, which is challenging due to the nature and limited knowledge about the human vision ~\cite{wyszecki2000color}. Multimedia compression standards such as JPEG and MPEG ~\cite{bhaskaran1997image} have been developed to compress visual multimedia such as images and videos to reduce the amount of data with minimum amount of distortion to the perceived output. One of the most fundamental ideas of visual multimedia compression is that human vision is much less sensitive to the information loss in color than the luminance.
This observation is utilized in image compression as a technique known as ``chroma subsampling'', which is simply decreasing the resolution of chrominance channels of an image by subsampling to reduce the image size. There are variants of chroma subsampling that only subsamples chrominance along horizontal axis (4:2:2) or both horizontal and vertical axes (4:2:0). Without further compression, (4:2:0) chroma subsampling reduces the size of an image effectively to half of its original size. Replacing the chroma components of the pixels in by neighboring chroma components does not yield visible artifacts. We employ this observation to derive a new type of adversarial attack based on spatial transformations in chroma channels of perceptual colorspaces. We apply spatial transformation only to the chroma components of input image while keeping the luminance component intact. Figure~\ref{fig:flowtochannels} shows the effect of a randomly initialized flow field applied to the luminance, chrominance and both set of channels. It is clear that spatial transformation in luminance channels causes visible distortions while chrominance only spatial transformations cause very subtle changes for human vision. This effect is much more highlighted when only the differences are observed after applying a flow field. Figure \ref{fig:diff} shows the absolute pixel difference from the initial image when the same flow field is applied to RGB, \(C_{b}C_{r}\) and a*b* channels, respectively.

\begin{figure*}[t]
    \centering
    \includegraphics[width=0.8\linewidth]{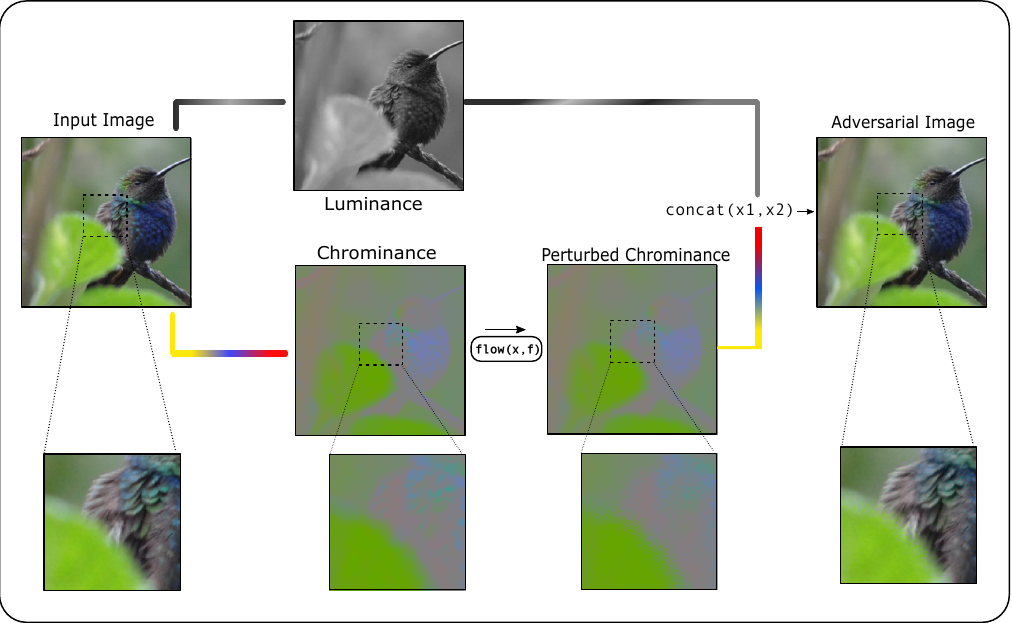}
    \caption{Visual illustration of the proposed adversarial example generation method with zoomed patches of each step at the bottom for better visual understanding. Luminance and chrominance channels are $Y$ and \(C_{b}C_{r}\) when \(YC_{b}C_{r}\) colorspace and L and \(a^*b^*\) when CIELAB colorspace is used. Visual representation of flow field f, optional subpixel restriction by \(\tanh\) and conversion of concatenated image back to RGB colorspace is omitted for brevity.}\label{fig:algorithm}
\end{figure*}
\section{Related Work}
Spatial transformations as methods for generating adversarial examples were first proposed in ~\cite{xiao2018spatially}, where it is shown that small displacements applied to input pixels can successfully fool a target network. However, using this method, even small displacements could cause visible distortions when the adjacent pixels are drifted towards different directions. As a remedy to this problem, use of  Total Variation~(TV) regularization~\cite{estrela2016total} was proposed. Application of TV regularization to the flow field pushes the neighboring displacement vectors to the same direction and, hence, produces smoother output. Also, Liu et al. utilized spatial transformations by applying flow field not directly to the original image, but to the latent representations from the the target network~\cite{liu2023dualflow} so that the generated adversarial example is more perceptually similar to the original image. Similarly, Jordan et al. \cite{jordan2019quantifying} combined spatial transformations with \(l_\infty\) bounded attacks to forge stronger attacks with better perceptual quality. Croce et al. argued adding noise to smooth areas of an image causes visible artifacts and proposed "hiding" the perturbations at the locations with high spatial variations such as edges and corners~\cite{croce2019sparse}. As seen from Figure \ref{fig:diff}, perturbations made with our method naturally occurs in the places with high variations since it is based on local spatial transforms. With a similar motivation, Luo et al.~\cite{luo2022frequency} proposed limiting the high frequency components of the adversarial perturbation to ensure perceptual similarity between adversarial and original examples, similar to the work of Duan et al.~\cite{duan2021advdrop}, which explored dropping the imperceptible information obtained by frequency domain conversion. 

Utilizing perceptual colorspaces and metrics for imperceptible adversarial example generation is investigated in several studies. Aksoy et al.~\cite{aksoy2019attack} investigated additive noise based attacks on chrominance channels in YUV colorspace, which is the analog counterpart of \(YC_{b}C_{r}\) space. Despite Pestana et al.~\cite{Pestana2020-hm} found that adversarial perturbations are more highlighted in luminance channels in terms of the magnitude, Aksoy et al. found that even suppressing the luminance perturbation, additive noise based attack on chrominance channels still successfully fool target networks, yet causes visible distortion. In our earlier work, we also explored spatial transformations to UV channels of YUV to generate imperceptible adversarial examples~\cite{aydin2019imperceptible} and we extend this work by exploring \(YC_{b}C_{r}\) space as well as perceptually uniform CIELAB space and measuring structured similarity metrics such as Structured Similarity Index Measure (SSIM)~\cite{wang2004image} and Multi-Scale Structured Similarity Index Measure (MS-SSIM)~\cite{wang2003multiscale} between benign images and adversarially generated images. Karli et al. \cite{karli2021improving} leveraged perceptual metric LPIPS~\cite{zhang2018unreasonable} to improve the quality of adversarial examples. Since LPIPS is a differentiable metric, they used gradient-based optimization to minimize LPIPS alongside the adversarial loss. Similarly, Zhao et al. replaced CIEDE2000 perceptual distance metric~\cite{zhao2020large} with \(\mathcal{L}_{p}\) norm constraint in Carlini \& Wagner attack to produce perceptually close adversarial examples. Chen et al.~\cite{chen2022imperceptible} utilized information preservation properties of Invertible Neural Networks~\cite{ardizzone2018analyzing} to minimize the adversarial perturbation with regards to perceptual distance metrics to generate imperceptible adversarial examples. 

Unlike these methods, the attack proposed in this paper does not rely on auxiliary losses or explicit perceptual distance terms in optimization process to produce examples with high perceptual quality. In addition, it does not require regularization, unlike spatial transformation based methods such as ~\cite{xiao2018spatially}, due to its intrinsic imperceptibility. It should be noted that the existing spatial transformation based methods, as well as our work, does not utilize limited degree of freedom transformations such as rotation, translation or scaling that can be formulated as a \(4\times4\) transformation matrix~\cite{jaderberg2015spatial}. In that formulation, the flow field \(f \in \mathbb{R}^{2\times H \times W}\) is calculated using the transformation matrix, where \(H, W\) are the height and the width of the image, respectively. Instead, we directly define and optimize flow field, where the number of parameters is equal to twice number of pixels in the input image since there is an x and y component for each pixel. Application and optimization of flow field is explained in the Section \ref{section:methodology}.

\section{Methodology}\label{section:methodology}
In this work, we address the problem of creating targeted adversarial examples for deep image recognition networks without adversarial perturbation being perceptible by human vision. To obtain this, we use a modified version of Spatially Transformed Adversarial Examples (stAdv)~\cite{xiao2018spatially} that perturbs the input image only in the channels that human vision is not sensitive to the spatial information loss. For this purpose, we use \(YC_{b}C_{r}\) and CIELAB colorspace representations of the input image. The proposed adversarial example generation method is as follows. Let \(x \in \mathbb{R}^{3\times H \times W}\) be the 3-channel input image. First, we randomly initialize a flow field \(f \in \mathbb{R}^{2\times H \times W}\) where a two-dimensional vector \(f^{(i,j)} = (\Delta i,\Delta j\)) for each pixel location \((i, j)\) in the adversarial image \(\mathbf{x}_{adv}\). Then, we apply the flow field to the benign image as explained below to obtain the adversarial image. Then, we feed the adversarial image to the target network and backpropagate the loss gradient to the flow field. Since the flow field application is a differentiable process, it can be optimized by stochastic gradient descent and variants such as Adam~\cite{kingma2015adam} or L-BFGS optimizers \cite{liu1989limited}. Hence, the target network could be any differentiable deep neural image recognition model. The optimization process is repeated until the attack is successful or when the maximum iteration count is reached. The process of generating adversarial image from the benign image is visualized in Figure \ref{fig:algorithm}.

\RestyleAlgo{ruled}
\begin{algorithm}[t]
    \caption{Adversarial example generation by spatial transformation in chrominance channels in a perceptual colorspace. }\label{alg1}
    \KwIn{   \(\mathit{x: Image}\)}
    \KwOut{   \(\mathit{x_{adv}: Image}\)}
    \KwData{
        \(target\_class: int\),
        \(model\),
        \(\kappa: float\),
        \(colorspace\),
        \(max\_iters: int\),
        \(is\_restricted: bool\),
    }
    \(f \gets initialize flow field f\)\;
    \(i \gets 0\);

    \While{\(i < max\_iters\)}{
    \If{\(colorspace == YC_{b}C_{r}\)}{
    \(x_{color} \gets to\_ycbcr(x)\)\;
    }
    \If{\(colorspace == CIELAB\)}{
        \(x_{color} \gets to\_lab(x)\)\;
    }
    \(x_{luma}, x_{chroma} \gets splitchannels(x_{color})\)\;
    \If{is\_restricted}{\(f \gets \tanh(f)\)}
    \(x_{chroma} \gets apply\_flow(x_{chroma}, f)\)\;
    \(x_{adv} \gets concat(x_{luma}, x_{chroma})\)\;
    \(x_{adv} \gets to\_rgb(x_{adv})\)\;
    \(adv\_scores \gets model(x_{adv})\)\;
    \(loss \gets loss\_fn(adv\_scores, target\_class, \kappa)\)\;
    \eIf{\(loss \leq \kappa\)}
    {\Return{\(x_{adv}\)}\;}
    {

        \(backprop(loss)\)\;
        \(update(f)\)\;
        \(i \gets i + 1\)\;
    }
    }
\end{algorithm}

\subsection{Application of flow field}
Flow field is applied to the benign image following the methodology in~\cite{xiao2018spatially}. For each pixel in the adversarial image \(\mathbf{x}_{adv}^{(i,j)}\), a pixel from location \((i + \Delta i, j + \Delta j)\) is sampled from the input image. Since the sampled location is not an integer value, 4 neighboring pixels around the location \((i + \Delta i, j + \Delta j)\) are (bilinear) interpolated \cite{zhou2016view} using Eq. \ref{eq:bilinear} where \(N(i + \Delta i, j + \Delta j)\) is the integer pixel positions around \((i + \Delta i, j + \Delta j)\). Bilinear interpolation also makes the method end-to-end differentiable, thus optimizable by gradient-based optimizers.

\begin{equation}
\label{eq:bilinear}
\mathbf{x}^{(i,j)}_{adv} = \sum_{\hat{i}, \hat{j} \in N(i + \Delta i, j + \Delta j)} \mathbf{x}^{(\hat{i},\hat{j})} (1 - |\hat{i} - (i + \Delta i)|) (1 - |\hat{j} - (j + \Delta j)|)
\end{equation}

Chroma subsampling effectively causes the same chroma values to be used by the neighboring pixels, and it is widely accepted to cause negligible changes to the images ~\cite{bhaskaran1997image}. Accordingly, to exploit this fact, we can impose a restriction to the flow-field \(f \in \mathbb{R}^{2\times H \times W}\) to keep its values in the range \((-1, 1)\) by using a hyperbolic tangent function, $\tanh(f)$, before applying the flow-field to the benign image.
This differentiable reparameterization~\cite{mordvintsev2018differentiable} of flow field constraints the flow field magnitude to be smaller than 1 without inhibiting end-to-end differentiability so that chrominance value of each pixel of the adversarial image \(x_{adv}\) is only affected by the value of the pixel of the same location in \(x\) and its neighboring pixels. 
\subsection{Colorspace conversion}
To make the adversarially perturbed images indistinguishable from their benign counterparts, the flow field is applied only to the channels that human vision is not very sensitive to~\cite{vorobyev2004ecology}. A perceptual colorspace is designed to approximate the way humans perceive color. It takes into account the characteristics of the human visual system and is used to represent colors in a way that is meaningful and intuitive to people.  Since widely used RGB colorspace is not designed to be a perceptual colorspace, even small spatial perturbations to any RGB channel creates visually distinguishable changes. Hence, we first convert the benign image to a perceptual colorspace such as \(YC_{b}C_{r}\) where human vision is not sensitive to the spatial perturbations in, which is \(C_{b}\) and \(C_{r}\) in \(YC_{b}C_{r}\), and \(a^*\) and \(b^*\) in CIELAB colorspace. Then, we apply the flow field only to the channels Cb and Cr in \(YC_{b}C_{r}\), and A and B in CIELAB colorspace.

\(YC_{b}C_{r}\) is a colorspace that is used in digital photography and visual media compression. In this space, luminance (brightness) and chrominance (color) is separated according to human visual perception. Y dimension of the space is the luminance information, or simply a grayscale representation of the image. \(C_b\) and \(C_r\) dimensions are the blue-difference and red-difference chroma components, respectively. The relation between RGB space and \(YC_{b}C_{r}\) space is modeled as Equation~\ref*{eq:eq1} and \ref*{eq:eq2}, which is a set of linear equations defined in ITU-T H.273~\cite{hamilton2004jpeg};
\begin{align}
    \label{eq:eq1}
    \begin{split}
        Y   & = 0.299 R+0.587 G+0.114 B                   \\
        C_b & = 128-(0.168736 R)-(0.331264 G)+(0.5 B)     \\
        C_r & = 128+(0.5 R)-(0.418688 G)-(0.081312 B)     \\
    \end{split}
\end{align}
\begin{align}
    \label{eq:eq2}
    \begin{split}
        R & = Y+1.402(C_{r}-128)                        \\
        G & = Y-0.34414(C_{b}-128)-0.71414(C_{r}-128)   \\
        B & = Y+1.772C_b-128
    \end{split}
\end{align}


CIELAB colorspace~\cite{schanda2007colorimetry}, defined by the International Commission on Illumination (CIE), has the following three components: L, \(a^*\) and \(b^*\). L is perceptual lightness where \(L = 0\) and \(L* = 100\) define a black and a white pixel, respectively, regardless of the \(a^*\) and \(b^*\) values. \(a^*\) and \(b^*\) dimensions are the chroma components. They are designed to be perceptually uniform where a numerical change in pixel value corresponds to a similar change in human perception ~\cite{mahy1992luminancevschroma}. Both chroma components are in the range \([-127, 127]\). Unlike \(YC_{b}C_{r}\), CIELAB space does not have a linear relationship with RGB space. In fact, conversion to an intermediary space CIEXYZ is needed to transform from RGB to CIELAB and there are different implementations of CIELAB conversion. We used the RGB to LAB implementation from Kornia library~\cite{riba2020kornia}, which assumes D65 illuminant and Observer 2, where D65 is a standardized illuminant that represents daylight with a correlated color temperature of 6500 K and the Observer 2 represents the human visual system's average response to color and is used as the standard observer in the CIELAB color space.

\begin{figure}[h]
    \includegraphics[width=0.24\linewidth]{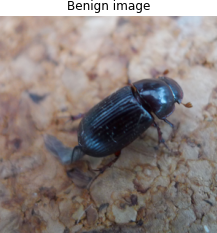}
    \includegraphics[width=0.24\linewidth]{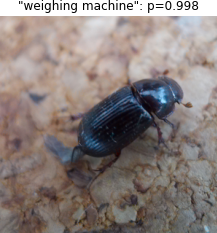}
    \includegraphics[width=0.24\linewidth]{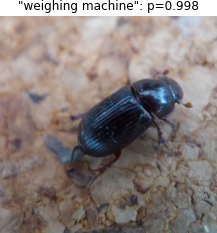}
    \includegraphics[width=0.24\linewidth]{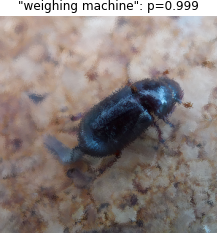}

    \includegraphics[width=0.24\linewidth]{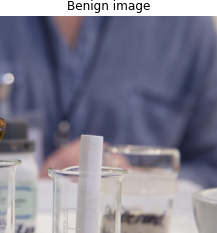}
    \includegraphics[width=0.24\linewidth]{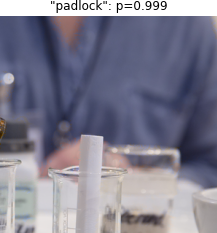}
    \includegraphics[width=0.24\linewidth]{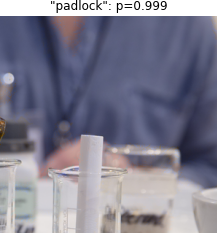}
    \includegraphics[width=0.24\linewidth]{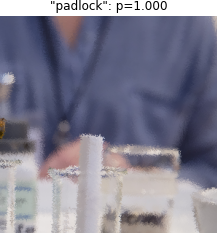}

    \includegraphics[width=0.24\linewidth]{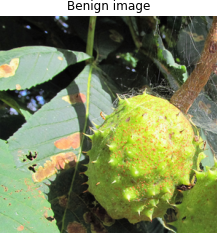}
    \includegraphics[width=0.24\linewidth]{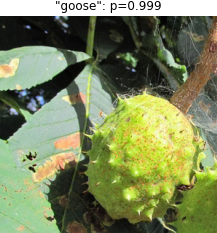}
    \includegraphics[width=0.24\linewidth]{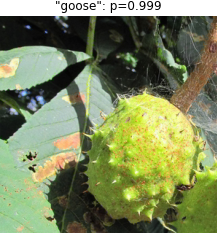}
    \includegraphics[width=0.24\linewidth]{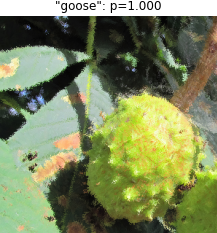}

    \includegraphics[width=0.24\linewidth]{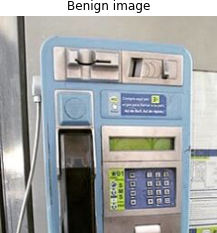}
    \includegraphics[width=0.24\linewidth]{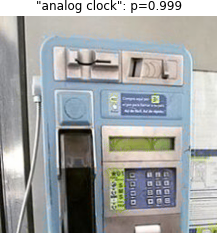}
    \includegraphics[width=0.24\linewidth]{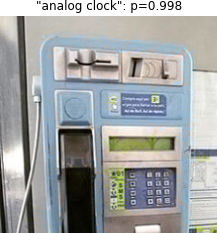}
    \includegraphics[width=0.24\linewidth]{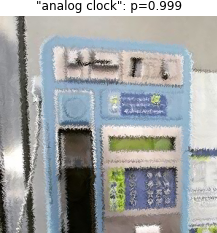}

    \includegraphics[width=0.24\linewidth]{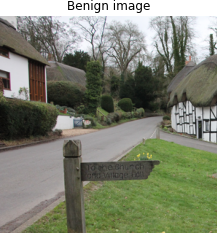}
    \includegraphics[width=0.24\linewidth]{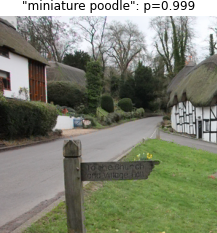}
    \includegraphics[width=0.24\linewidth]{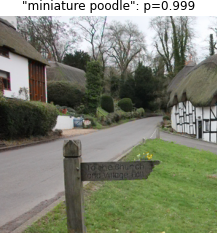}
    \includegraphics[width=0.24\linewidth]{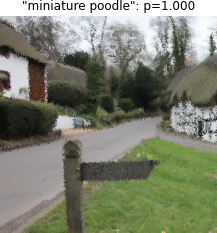}

    \includegraphics[width=0.24\linewidth]{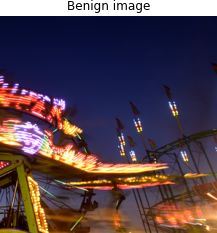}
    \includegraphics[width=0.24\linewidth]{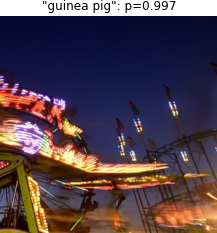}
    \includegraphics[width=0.24\linewidth]{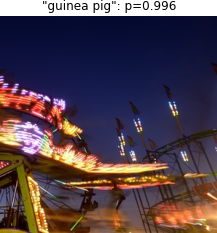}
    \includegraphics[width=0.24\linewidth]{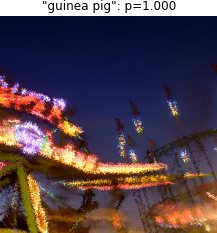}

    \caption{Examples from the dataset and adversarial examples generated with their target class probabilities. From left to right; Benign image, adversarial image generated by perturbing CbCr, a*b* and RGB channels, respectively. }\label{fig:visualprob}
\end{figure}
\section{Experimental Evaluation}
We conducted our experiments in a white-box setup where the gradients are fully available. Experiments have been done in a targeted attack setting with the dataset provided targets. We optimized using Adam ~\cite{kingma2015adam} with the default settings and used Carlini \& Wagner loss ~\cite{carlini2017towards} with confidence margin \(\kappa \in \left\{ 0, 10 \right\}\).


We used the dataset and the provided model from NIPS 2017 Competition on Adversarial Attacks and Defenses~\cite{kurakin2018adversarial} to evaluate our method. NIPS 2017 dataset is a collection of 1000 images curated by Google Brain with the resolution of \(299 \times 299\) with their corresponding true and target classes from Imagenet~\cite{deng2009imagenet} dataset. Alongside the dataset, an Imagenet trained Inception-v3~\cite{szegedy2016rethinking} model is provided. 

We compared the success rate of our attack in CIELAB and \(YC_{b}C_{r}\) against stAdv in both restricted and unrestricted settings. An attack is considered successful if the Carlini \& Wagner loss is less than \(-\kappa\). We did not use the smoothness regularization term in stAdv for a fair comparison.
\begin{table}[t]
    \caption{Attack success rates with \(\kappa = 0\) and \(\kappa = 10\) in not restricted and subpixel restricted settings for RGB, \(a^*b^*\) and \(C_{b}C_{r}\) attacks. }

    \begin{tabularx}{\linewidth}{ X  X  X  X }
        \toprule
                                           & RGB                   & \(C_{b}C_{r}\)        & \(a^{*}b^{*}\)        \\
        \hline
        \multicolumn{4}{c}{Not Restricted}                                                                         \\
        \midrule
        $\kappa$ = 0\newline $\kappa$ = 10 & 100\%\newline 100\%   & 95.0\%\newline 83.8\% & 95.7\%\newline 87.3\% \\
        \hline
        \multicolumn{4}{c}{Restricted to Subpixel}                                                                 \\
        \midrule
        $\kappa$ = 0\newline $\kappa$ = 10 & 99.8\%\newline 99.7\% & 86.1\%\newline 47.0\% & 89.2\%\newline 53.2\% \\
        \bottomrule
    \end{tabularx}\label{table:foolingrate}
\end{table}

\begin{table}[t]
    \caption{Average amount of distortion required to fool the target network with very high confidence (\(\kappa=10\)) in not restricted and subpixel restricted settings.}
    \label{table:perceptualmetrics}
    \begin{tabularx}{\linewidth}{ X  X  X  X }
        \toprule

                                           & RGB                               & \(C_{b}C_{r}\)                                             & \(a^*b^*\)                                \\
        \hline
        \multicolumn{4}{c}{Not Restricted}                                                                                                                                              \\
        \midrule
        LPIPS\newline SSIM\newline MS-SSIM & 0.327\newline 0.321\newline 0.164 & \textbf{0.019}\newline \textbf{0.067}\newline 0.017        & 0.022\newline 0.070\newline\textbf{0.016} \\
        \hline
        \multicolumn{4}{c}{Restricted to Subpixel}                                                                                                                                      \\
        \midrule
        LPIPS\newline SSIM\newline MS-SSIM & 0.222\newline 0.220\newline 0.037 & \textbf{0.012}\newline\textbf{0.050}\newline\textbf{0.011} & 0.014\newline 0.056\newline 0.013         \\
        \bottomrule
    \end{tabularx}
\end{table}
\subsection{Analysis of the Results}
Figure~\ref{fig:visualprob} shows the original images alongside with the adversarial images generated (with \(\kappa = 10\)) by attacking in \(a^*b^*\), \(C_{b}C_{r}\) and RGB spaces. As can be observed from these images, perceptual distortions are much less pronounced for chrominance-only attacks. Attacking in RGB domain, which is the default approach in the literature, results in modification of the luminance channels, leading to much more visible artifacts.

Table~\ref*{table:foolingrate} shows the attack success rates for attacks on different colorspaces. The results show that, adversarial images generated by attacks exclusively targeting the chrominance channels can fool the network with a high probability as well. On the other hand, they are less effective when restricted to operate in a subpixel-only setting. The fooling rate of a*b* attacks are slightly higher than \(C_bC_r\) attacks. We argue that this is due to many examples in the dataset being chroma subsampled in \(YC_bC_r\) space, as an indirect effect of image compression, restricting the search space for \(C_bC_r\) attacks.

We measured the amount of distortion required to generate confident (\(\kappa = 10\)) adversarial examples with the following perceptual metrics: Learned Perceptual Image Patch Similarity~(LPIPS) ~\cite{zhang2018unreasonable}, Structured Similarity Index~(SSIM) ~\cite{wang2004image} and Multi-Scale SSIM~(MS-SSIM) ~\cite{wang2003multiscale}. Table~\ref{table:perceptualmetrics} shows the average results over the successful attacks for each perturbation mode in terms of these metrics. Since SSIM and MS-SSIM are similarity metrics, values of \(1-\)SSIM and \(1-\)MS-SSIM are provided. Hence, for all metrics, lower values are better. According to these results, colorspace restricted attacks have much better scores in terms of perceptual metrics compared to RGB attacks, implying that there is significantly less perceptual difference between benign and adversarial examples. While \(C_bC_r\) attacks generally produce better images in terms of perceptual quality metrics than a*b* attacks, the difference is relatively low.

\begin{figure}[t]
    \includegraphics[width=0.328\linewidth]{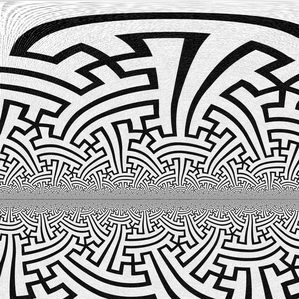}
    \includegraphics[width=0.328\linewidth]{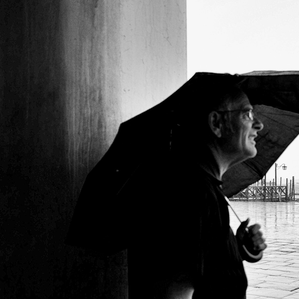}
    \includegraphics[width=0.328\linewidth]{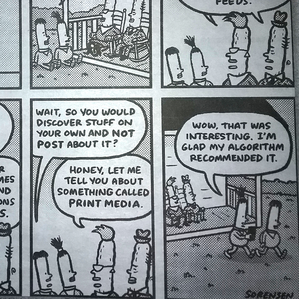}
    \includegraphics[width=0.328\linewidth]{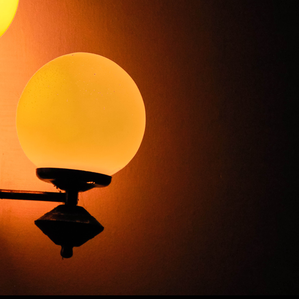}
    \includegraphics[width=0.328\linewidth]{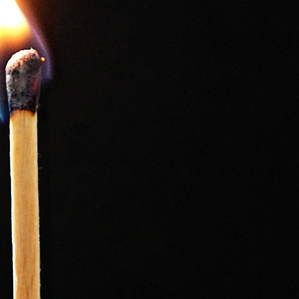}
    \includegraphics[width=0.328\linewidth]{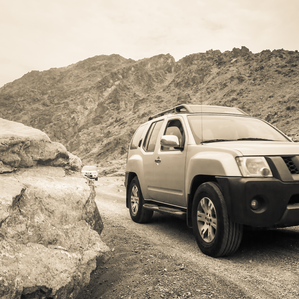}
    \includegraphics[width=0.328\linewidth]{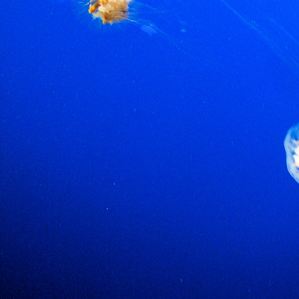}
    \includegraphics[width=0.328\linewidth]{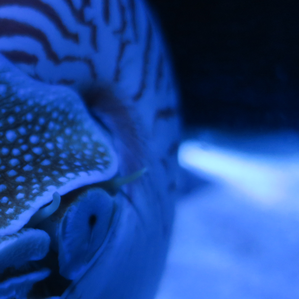}
    \includegraphics[width=0.328\linewidth]{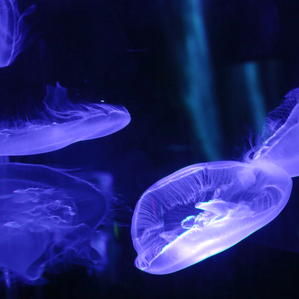}
    \caption{Examples from the dataset that our method fails to generate successful adversarial examples from in both \(YC_{b}C_{r}\) and CIELAB spaces, sorted from top bottom by colorfulness amount.}\label{fig:fails}
\end{figure}

\subsection{Analysis of Failure Cases}
Experimental results show that there are two main restrictions of the proposed method: out of gamut values in the chrominance channels emerging during optimization leading to visible artifacts and failing to generate adversarial images when the original image has limited colorfulness.

\textbf{Failed Attacks on Less Colorful Images}: Results in Table~\ref{table:foolingrate}, show that the attack success rate does not reach 100\% when spatial transform attack is restricted to chrominance channels.  
This implies that the chrominance based attacks fail for a number of images in the dataset. Examples of such images are provided in Figure~\ref{fig:fails}. We observed that these  particular images are either monochromatic examples or have a uniform color pattern, for which spatial transformation in a neighborhood lead to little change.

\begin{figure}[t]

    \begin{center}
        \includegraphics[width=0.92\linewidth]{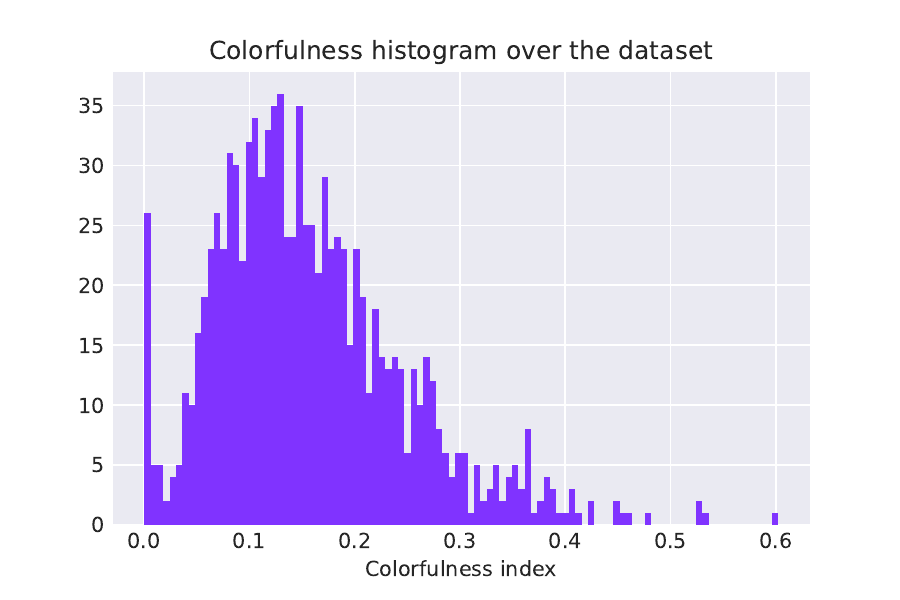}
    \end{center}
    \caption{Colorfulness index histogram over NIPS2017 dataset.}\label{fig:hist}
\end{figure}

\begin{figure}[t]
    \begin{center}
        \includegraphics[width=0.92\linewidth]{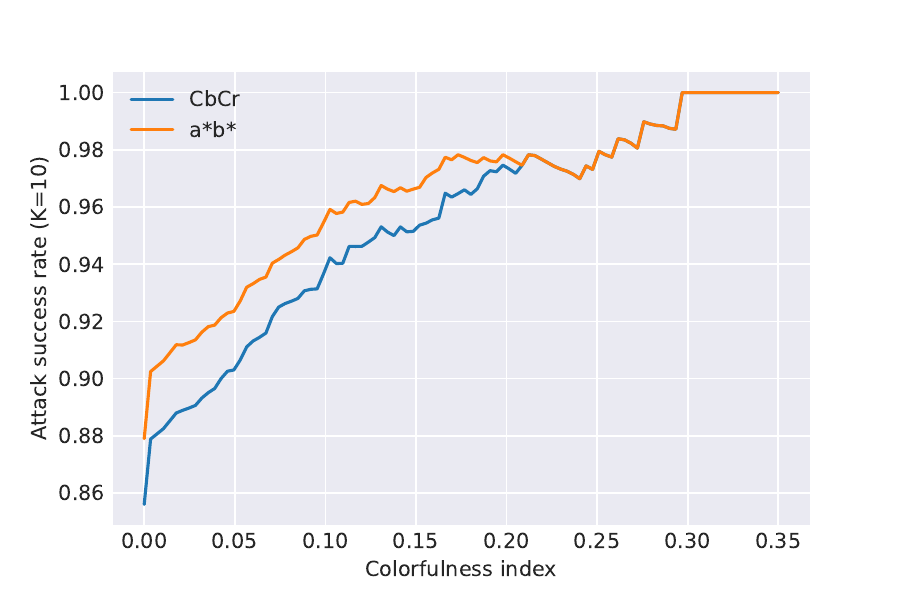}
    \end{center}
    \caption{Attack success rate analysis with regards to colorfulness index with \(\kappa=10\) on \(CbCr\) and a*b* channels. Images having colorfulness index less than the $x$ axis value are excluded in calculation of the success rate. Note that both colorspaces attain very close success rates after around colorfulness index 0.2.} \label{fig:plots}
\end{figure}

To analyze the effect of colorfulness on the attack performance, we calculated the colorfulness index histogram of the images in the dataset (Figure \ref{fig:hist}) . We found that 3.2\% of the dataset consists of grayscale images, for which our method would not be able to make any changes to the input image, inevitably resulting in a failed attack. Figure \ref{fig:plots} shows the attack success rate using the subsets where colorfulness is lower-limited by filtering out examples having colorfulness index less than the $x$ axis value. Although a*b* attacks are slightly more successful than \(C_bC_r\) in the low colorfulness regime \((\leq 0.2)\), they have the same success rate of the attacks over higher colorfulness.

\begin{figure}[t]
    \includegraphics[width=0.495\linewidth]{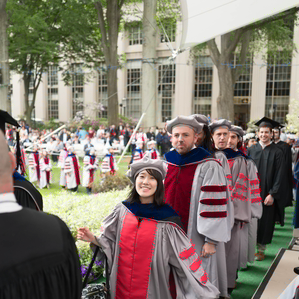}
    \includegraphics[width=0.495\linewidth]{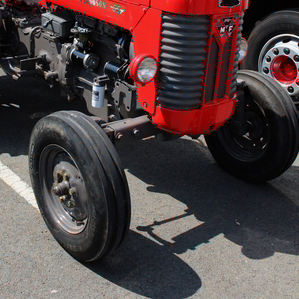}
    \caption{Examples of visible clipping artifacts of out-of-gamut pixels caused by spatial transform around red-gray borders. Flow magnitude has been scaled up to highlight the visible effects for illustration.}\label{fig:outofgamut}
\end{figure}

\textbf{Out of Gamut Values}: Modifying the chrominance channels in \(YC_{b}C_{r}\) and CIELAB spaces may lead to improper values on individual RGB channels. This is also common in widely used chroma subsampling and mitigating this issue is an open research topic~\cite{chan2008toward}. In our work, we clip the reconstructed RGB to the valid range and feed the target network with the clipped image at each iteration to prevent further change in the pixel values out of the gamut. Clipping also zeroes out the gradient and prevents further updates in gradient-based optimization. However, we found that it still causes visible artifacts in the adversarial image, especially around the borders between red and gray tones. Figure~\ref{fig:outofgamut} shows two examples where spatial transformation in red-gray borders yield out of gamut pixels and clipping the values still causes visible artifacts since clipping in RGB space effectively changes the values of luminance channels.




\section{Discussion}\label{section:discussion}
As it can be seen by Figure~\ref{fig:fails}, the input images that our method fails are generally grayscale or monochromatic images, which prevents chrominance spatial transforms from changing the pixel values due to the low magnitude of chrominance channel values. In addition, input images having a very limited local color variation negatively affect the performance by limiting the potential search space. We observed that there is a significant drop in the success rate with the setup confidence margin \(\kappa=10\) if the attack is restricted to subpixel changes in comparison to the unrestricted attacks. We argue that this performance drop is arising from the fact that the most examples are already JPEG compressed, which means chroma subsampling is applied to the benign examples, which restricts the subpixel restricted search space by dramatically reducing the local chrominance variation. This leads to the observation that chroma subsampling could be an effective defense method against our attack. Moreover, the search space is further restricted in JPEG compressed images as the quantization step of JPEG compression attenuates high frequency information, especially in the chrominance channels. Nonetheless, we observed adversarial examples generated by spatial transforms in chrominance channels of perceptual colorspaces obtain competitive fooling rates without making perceptible changes to the image. This observation provides further evidence for the hypothesis that representation of deep neural networks does not necessarily align with human vision ~\cite{geirhos2018imagenet}.

\section{Conclusions}
Adopting the techniques used in multimedia compression and using the idea that pixel shifts in a constrained neighborhood are hard to notice, we designed a method that applies local spatial transformations to chrominance channels of perceptual colorspaces. The proposed method results in adversarial images having imperceptible distortions without requiring any regularization. In addition to obtaining competitive fooling rates, restricting magnitude of the spatial transformations still yields successful attacks, when there is sufficient amounts of local chrominance variation in the input image.

In addition to the perceptual colorspaces investigated in this work, other perceptual colorspaces such as CIELUV, HSLuv and CIEXYZ ~\cite{schanda2007colorimetry} can also be utilized to create imperceptible adversarial examples. Out of gamut values at borders with red pixels may result in visible artifacts during the adversarial image generation and preventing such out-of-gamut values would result in better quality adversarial images.  
While our method does not require optimizing using a visual quality metric, it can be utilized along with our method to obtain a better visual quality.
\section*{Acknowledgements}
This work has been funded by The Scientific and Technological Research Council of Turkey, ARDEB 1001 Research Projects Programme project no: 120E093

\bibliographystyle{elsarticle-num}
\bibliography{mybibfile}

\end{document}